\documentclass[letterpaper, 10 pt, conference]{ieeeconf}  

\usepackage{graphicx}
\usepackage{stfloats}
\usepackage{float}
\usepackage{booktabs}
\IEEEoverridecommandlockouts                              

\usepackage{color} 
\usepackage{lipsum} 
\usepackage[backend=bibtex, style=numeric-comp, sorting=none]{biblatex}
\title{\LARGE \bf
Data-Driven Shape Sensing in Continuum Manipulators via Sliding Resistive Flex Sensors
}

\author{Chenhan Zhang$^{1, 2*}$, Shaopeng Jiang$^{2*}$, Heyun Wang$^{1, 2}$, Joshua Liu$^{2}$, \\ Amit Jain$^{2, 3}$, and Mehran Armand$^{1, 2, 3}$, \textit{Member}, \textit{IEEE}
\thanks{*The first two authors contributed equally to this work.}
\thanks{*This work has been submitted to the IEEE for possible publication. Copyright may be transferred without notice, after which this version may no longer be accessible.}
\thanks{$^{1}$Department of Mechanical Engineering, Johns Hopkins University, Baltimore, MD, USA. }%
\thanks{$^{2}$Biomechanical- and Image-Guided Systems (BIGSS) Laboratory within LCSR, Whiting School of Engineering, Johns Hopkins University, Baltimore, MD, USA}
\thanks{$^{3}$Department of Orthopaedic Surgery, Johns Hopkins School of Medicine, Baltimore, MD, USA}
\thanks{Corresponding author: Joshua Liu, \texttt{jsliu@jhu.edu}}
}

\begin{document}

\maketitle
\thispagestyle{empty}
\pagestyle{empty}

\begin{abstract}

We introduce a novel shape-sensing method using Resistive Flex Sensors (RFS) embedded in cable-driven Continuum Dexterous Manipulators (CDMs). The RFS is predominantly sensitive to deformation rather than direct forces, making it a distinctive tool for shape sensing. The RFS unit we designed is a considerably less expensive and robust alternative, offering comparable accuracy and real-time performance to existing shape sensing methods used for the CDMs proposed for minimally-invasive surgery. Our design allows the RFS to move along and inside the CDM conforming to its curvature, offering the ability to capture resistance metrics from various bending positions without the need for elaborate sensor setups. The RFS unit is calibrated using an overhead camera and a ResNet machine learning framework. Experiments using a 3D printed prototype of the CDM achieved an average shape estimation error of 0.968 mm with a standard error of  0.275 mm. The response time of the model was approximately 1.16 ms, making real-time shape sensing feasible. While this preliminary study successfully showed the feasibility of our approach for C-shape CDM deformations with non-constant curvatures, we are currently extending the results to show the feasibility for adapting to more complex CDM configurations such as S-shape created in obstructed environments or in presence of the external forces. 


\end{abstract}

\section{INTRODUCTION}
\par The Continuum Dexterity Manipulator (CDM) is a robotic structure with a dual backbone kinematic model comprising rigid links and joints, making it particularly suitable for robotic surgical applications. Its high manipulability, accessibility, and ease of miniaturization \cite{Segreti2012-fb,Shi2017-nm,Burgner-Kahrs2015-rs,Dupont2022-qh,Gao2022-fu,Abushagur2014-iv,10324929,Ma2021-uz}. These advantages facilitate CDMs in navigating intricate spaces and interacting with obstacles, making them ideal for Minimally Invasive Surgery (MIS) and endoscopic surgery \cite{Shi2017-nm}, \cite{Yip2014-mm}, \cite{Cao2023-cn}. However, CDM compliance poses challenges in real-time position sensing and shape estimation for each robot joint.
\par  The literature presents several primary approaches to shape sensing methods, including vision-based, model-based, and sensor-based techniques. Vision-based motion capture systems are known for their high accuracy. However, they are not suitable for the obstructed field of view in which CDMs operate\cite{Russo2023-uv,}.  Fluoroscopy overcomes the limitations of visual obstructions, but relies heavily on high image quality, and high frame rates pose challenges for real-time accurate MIS procedures  \cite{Gao2022-fu} .  Model-based methods have been developed based on various assumptions and the complex kinematic and dynamic characteristics of CDMs. Despite their potential, the extended computational time required for real-time applications and their susceptibility to environmental disturbances limit their generalizability\cite{Yip2014-mm,Gao2017-aw}. The Fiber Bragg Grating (FBG) shape sensor is famous for its high precision, small size, and bio-compatibility \cite{Shi2017-nm}, \cite{Amirkhani2023-qk}. but broader applications of FBG are often limited by the high cost and bulky nature of Bragg wavelength monitors \cite{Flores-Bravo2022-kt}. The Electromagnetic (EM) tracking systems can also be miniaturized to provide precise tracking on continuum robots \cite{Lugez2015-hf}. However, The EM tracking systems are facing challenges due to their sensitivity to electronic and metal instruments used during surgery \cite{Franz2014-lo}.

\par Resistive Flex Sensors (RFS) are extensively utilized in robotics and biomedical devices due to their linearity and sensitivity to angular displacements. This sensor offers several advantages, including durability, affordability, robustness, and resistance to electromagnetic interference \cite{Saggio2016-qc}.  The RFS has been widely utilized for soft grasping manipulator tip sensing with high accuracy \cite{Elgeneidy2018-so}. However, traditional implementations typically fix the sensor rigidly to the device, thereby only enabling point configuration measurements at predetermined positions. For continuum manipulators with varying curvatures, achieving global shape measurement requires an array of flex sensors at multiple joint positions. This not only increases system redundancy but also introduces added intricacies to the design and control mechanisms, amplifying both mechanical and control system challenges.

\begin{figure}[H]
   \centering
    \includegraphics[width=0.42\textwidth]{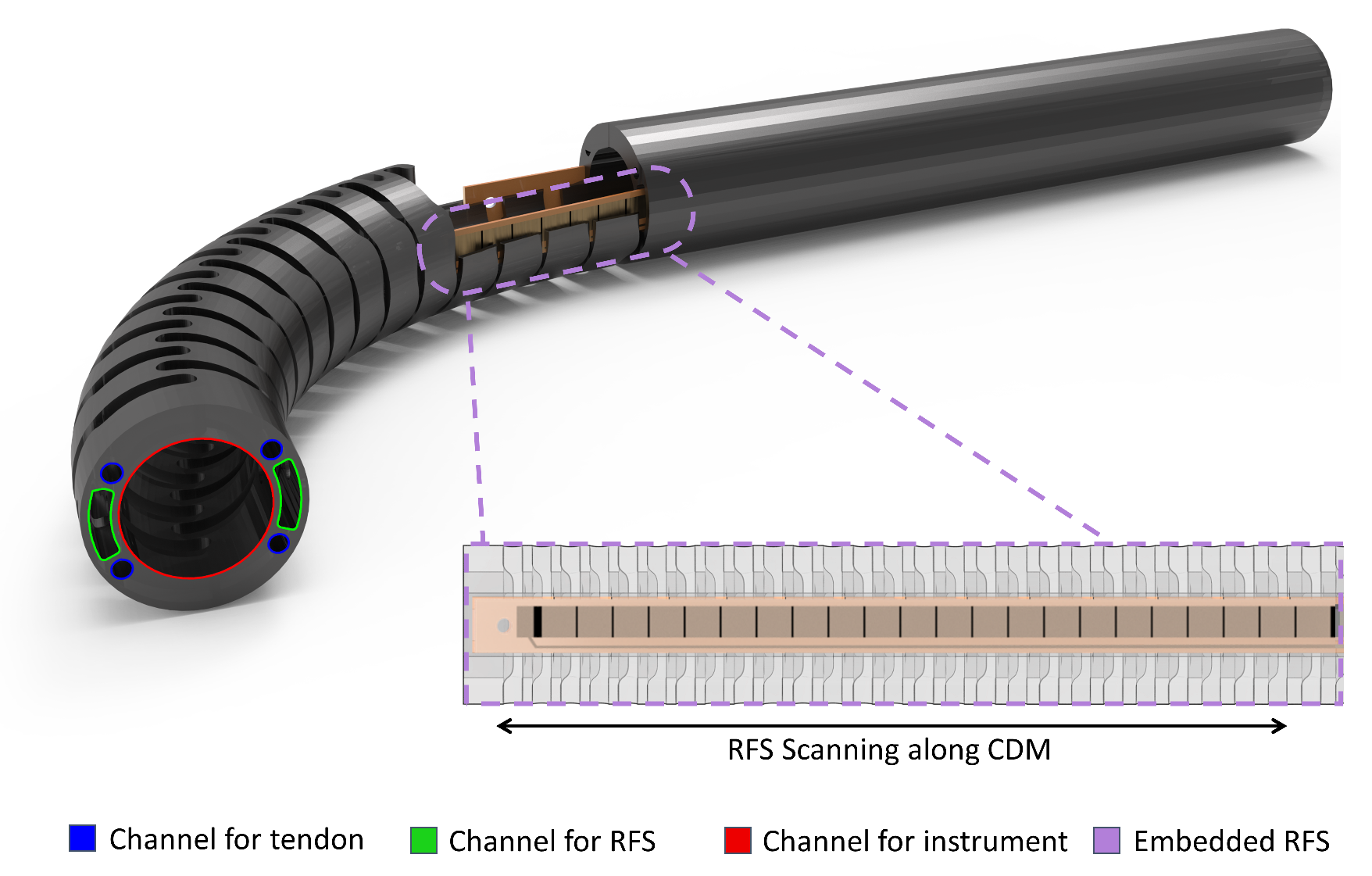}
    \caption{Overview of CDM and RFS unit.}
    \label{fig:cad}
\end{figure}

\par In this paper, we introduce a novel sliding RFS mechanism as a cost-effective solution to sense and reconstruct the shape of CDMs. Specifically designed for cable-driven CDMs with non-constant curvature configurations, this innovative mechanical design ensures real-time position detection across all selected joints. In place of conventional physical models, our approach leverages data-driven techniques for sensor calibration, ensuring accurate reconstruction of the CDM joint positions. We demonstrate that our RFS-based method can proficiently scan and measure diverse C-shaped configurations, especially in large bending scenarios where constructing precise physical models proves challenging. To the best of our knowledge, this marks the first application in measuring and reconstructing the shape of a continuum robot with dynamically sliding flex sensors through a data-driven approach.


\section{Method}

\subsection{Design and Fabrication of RFS Unit}

\begin{figure*}[t]
    \centering
    \includegraphics[width=0.85\textwidth]{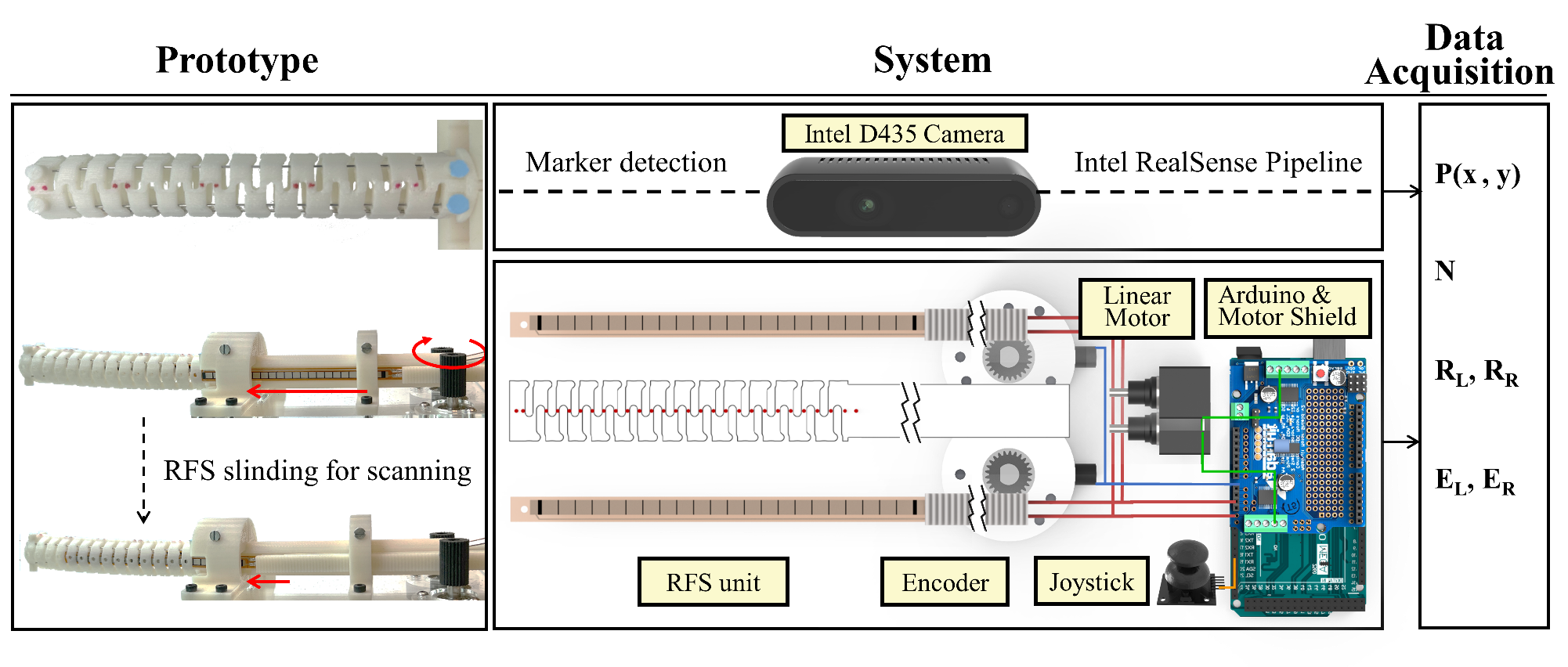}
    \caption{Workflow of the overall system.}
    \label{fig:control}
\end{figure*}

The RFS embedded in the CDM is the primary unit for shape sensing, as shown in Fig. 1.
For this feasibility study, we purchased a prefabricated RFS with a given measuring range (95.25 mm) and dimension (112.24 mm length, 6.35 mm width, 0.5 mm thickness). We designed a CDM prototype and a mechatronics testing system that corresponded to RFS features. This CDM prototype is scaled up three times from the typical CDMs proposed by our lab to be used in minimnally-invasive orthopedic procedures \cite{Sefati2022-fw,Alambeigi2019-sr,Ma2021-uz}. It was fabricated using U-Print F370 (AdvancedTek, MN), with a layer thickness of 0.1mm. It is composed of PC-ABS and consists of 26 notches that can be approximated by flexible joints, The outer diameter of the CDM prototype is 18 mm, and its inner diameter is 12 mm. The total length of the steerable portion of the CDM prototype is 102.14 mm.  The CDM offers degrees of freedom in one plane for flexibility and, in order to maintain stability during orthopedic surgical procedures, is rigid in the plane orthogonal to the plane that it flexes in. The CDM prototype includes seven thru channels. The surgical instrument channel is the largest with a 12 mm diameter. Four centrally symmetrical channels with 1.8 mm diameter, within the walls of the CDM, are used to guide the 0.65 mm diameter driven cable. The RFS units slide smoothly in two symmetrical channels within the walls of the CDM with 1.8 mm width.  To improve calibration and accuracy in identifying joint positions using the camera, we included additional design features. First, there is a 3 mm-wide and 1 mm-tall extrusion column running along the centerline of the entire CDM, designed to facilitate marker painting. Additionally, there are two pairs of large horizontal circles extruded at the tip and base of the CDM, serving as the reference points for all marker coordinate system conversions and tip bending angle estimations.

\par The RFS scanning mechanism consists of two gear racks in which their displacements are measured by two encoders. Each rack gear contains one RFS sliding inside the CDM. The default measuring range of the RFS is from 0 to 180 degrees, covering 95.25 mm active length along the CDM. Although the easiest way to use the sensor is to measure the bending and position of the tip of the CDM, the scanning mechanism enables the sensor to slide and scan all joints within the channel of the CDM. Therefore, incorporating the encoder, the sensor can measure the bending of the whole CDM body instead of the tip only. Due to the asymmetry of the RFS when bending in opposing directions, we insert one RFSs on each side of the CDM to cover the entire configuration space of the CDM ranging from -90 to 90 degrees.

\subsection{System Design and Integration}

This mechatronic system in Fig. 2 consists of two RFSs, two optical rotary encoders, two large torque linear motors,  one joystick, one Arduino Mega 2560, one Arduino Uno, one Adafruit motor shield, and one RGB camera (Intel RealSense D435). The camera is positioned above the CDM for marker identification and position detection. The remainder of the components are integrated into an Arduino Mega 2560 that communicates with the host computer via serial communication.
\begin{figure}[t]
   \centering
    \includegraphics[width=0.48\textwidth]{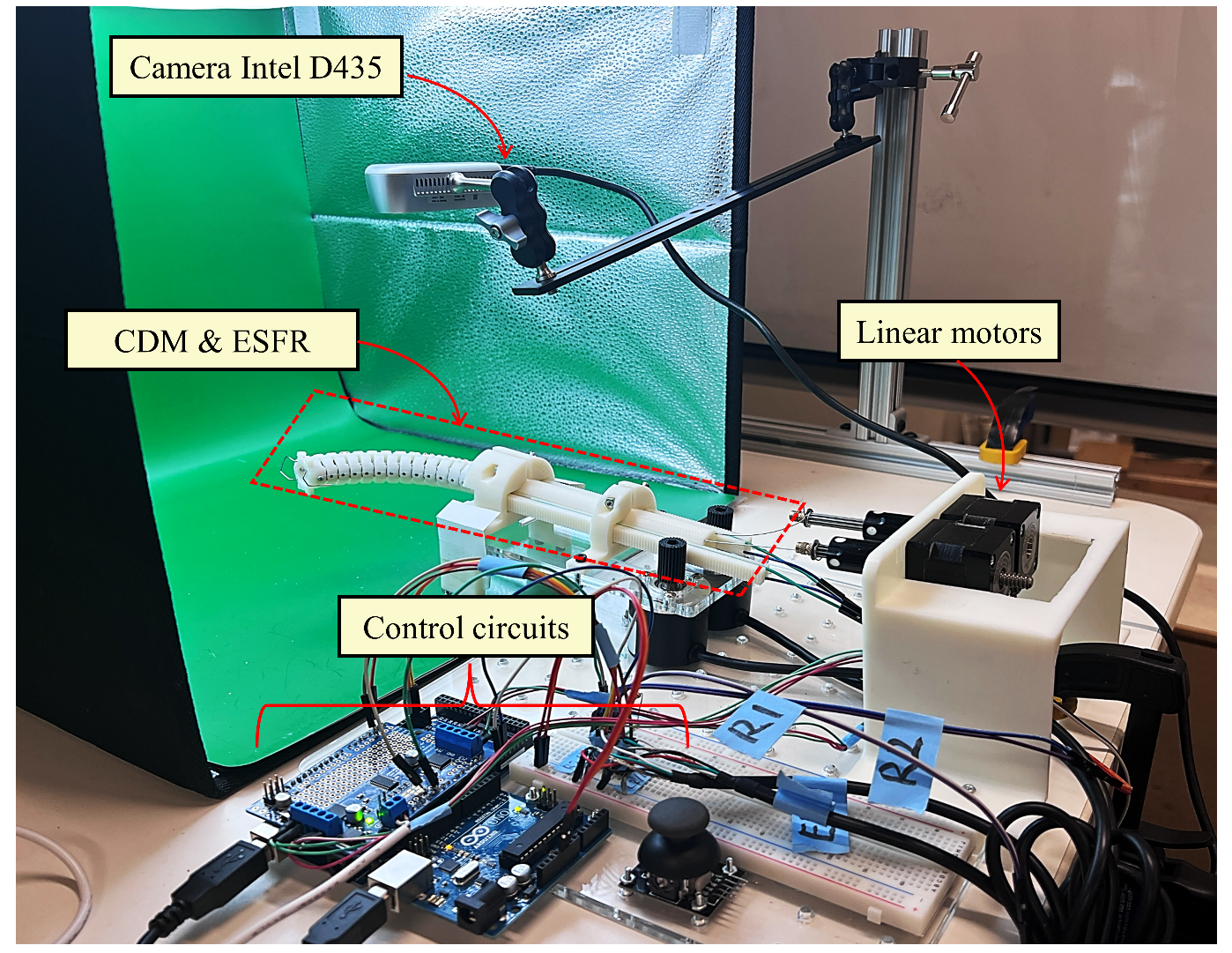}
    \caption{Experiment setup.}
    \label{fig:sys}
\end{figure}
Two RFSs are soldered to jump wires that are snapped into the rack and connected to the analog pins on Arduino Mega. A corresponding 10 k-ohm voltage division resistor on the breadboard will linearize the signal, improve its stability, and enhance signal resolution for each RFS. The two gear racks slide inside the slots along the CDM, allowing the encoders to detect the tip positions of the RFS. The encoders were connected to the Interrupt ReQuest (IRQ) pins via the Interrupt Service Routine (ISR) for real-time response and time consistency. \\

\par The joystick provided two analog signals for its X and Y directions, which were combined to control two linear step motor motions for bending and pretension of the cable-driven CDM. In addition, there were two microcontrollers, with the Arduino Mega 2560 packed with an Adafruit motor shield for CDM control and to collect all sensor data. The Arduino Uno is solely utilized to provide stable 5V current individually to RFSs in order to enhance signal quality.

\section{Results and Discussion}

\subsection{Experimental Setup}

The experimental setup included the actuation system, the CDM with embedded RFS units, the control circuit, and the camera system, as shown in Fig. \ref{fig:sys}. The actuation system is composed of a pair of stepper motor linear actuators (5.7 W, 35H4, Haydon Switch $\&$ Instrument Inc., Waterbury, CT, USA) with a pre-tensioned high-strength metal string that serves as the actuation tendon for the CDM. Before bending the CDM, the flexible sensor tip (FS-L-095, Spectra Symbol, Salt Lake City, UT) in RFS unit was slided into the CDM wall. Two rack gears were meshed with the spurs connected to the shafts of the two optical encoders (Rotary, LPD3806, China). The control circuit consisted of an Arduino UNO Board for constant power supply to RFS, and an Arduino MEGA board for actuation and data acquisition of the sensors and encoders. To make a uniform and consistent lighting, the RealSense D435 camera (Intel Corp., Santa Clara, CA, USA) was placed inside a photo light box (PU5041B, PULUZ, China) with a green background. The camera is mounted 15 cm above the CDM bending area to capture RGB images of all markers. The sampling frequency of the system is 20 Hz.
\begin{figure}[t]
   \centering
    \includegraphics[width=0.485\textwidth]{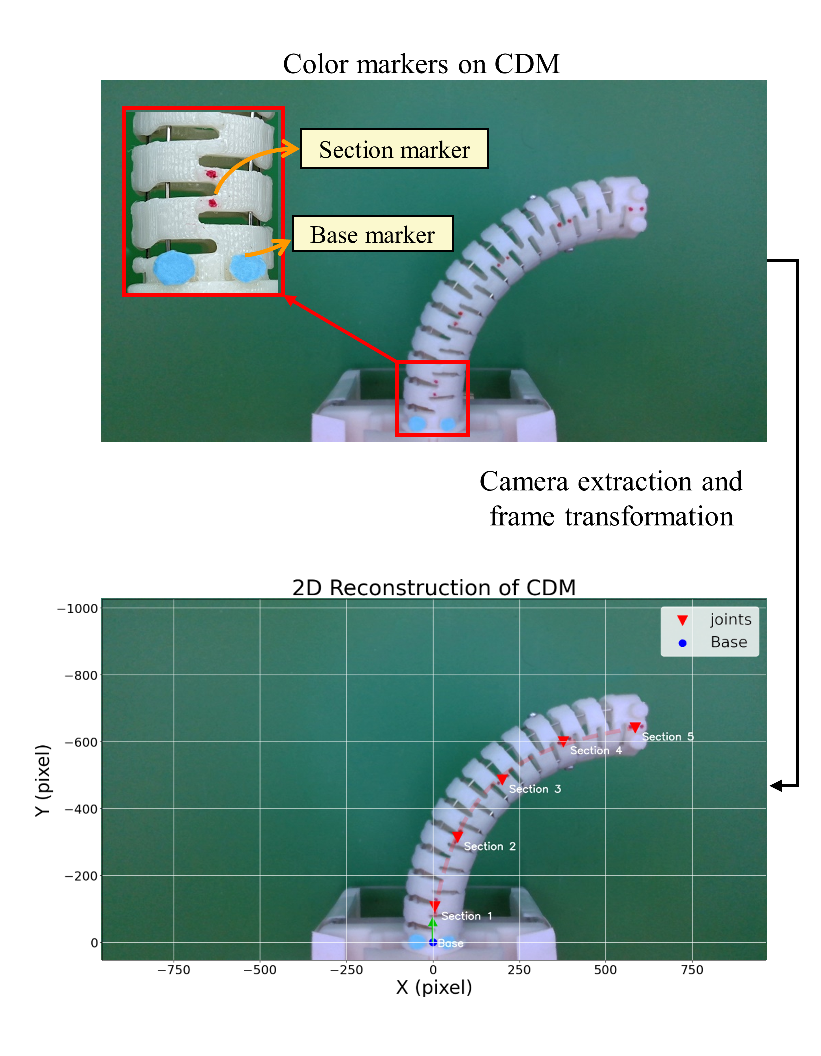}
    \caption{Camera feature extraction and reconstruction.}
    \label{fig:cam}
\end{figure}

\subsection{Image Processing}

An Intel RealSense D435 Camera with a resolution of $1920\times1080$ was utilized to detect five pairs of red markers situated at the center of specific CDM joints, in addition to a pair of blue markers that denote the base location and orientation of the CDM. The calibration of the camera's intrinsic and extrinsic parameters was performed using a checkerboard pattern, facilitated by the OpenCV Python Library. After calibration, the average reprojection error was found to be 0.13 pixels, and the scaling factor from the camera frame to real-world frame was found to be 10.37 pixels/mm. This was based on the accuracy with which the checkerboard pattern's corners were determined. An algorithm was formulated using an HSV filter, streamlining the efficient segmentation of markers based on their color. This enables the precise determination and 2D reconstruction of the centroids of the markers in terms of their 2D coordinates, as illustrated in Fig. \ref{fig:cam}. The camera-derived data is subsequently used for calibration of RFS sensing unit, as well as validation of the CDM shape reconstruction model as a ground truth.

\begin{figure}[t]
   \centering
    \includegraphics[width=0.4\textwidth]{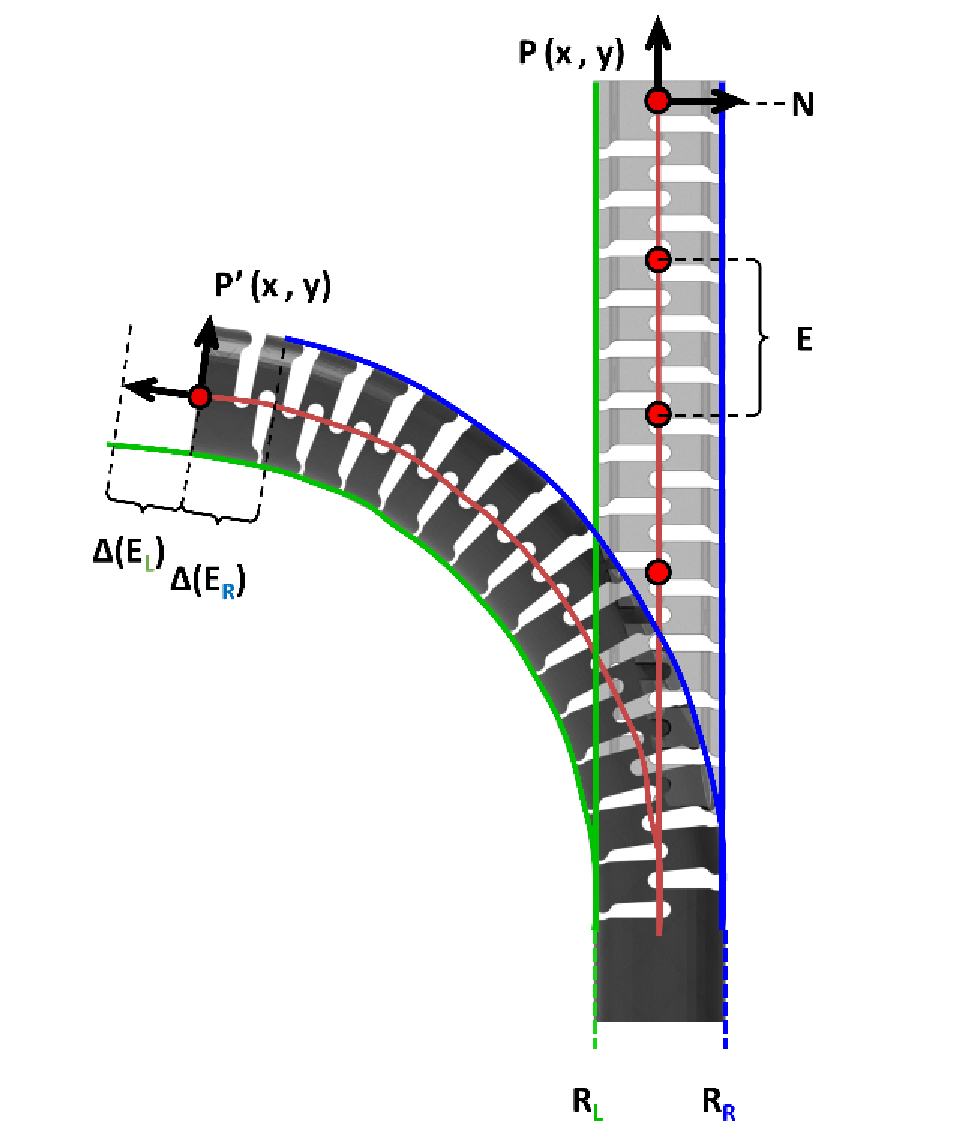}
    \caption{Schematic view of CDM and RFS.}
    \label{fig:variable}
\end{figure}
\subsection{Data Collection in 2D Free Space}

In an unconstrained 2D space, a series of five experimental trials were conducted for both positive and negative deflections of one distinct CDM section. Each trial comprised two primary cycles: the bending phase and the straightening phase. The bending cycle originated from a state of zero deflection where the CDM was straight. Subsequently, based on the direction of deflection, the corresponding actuating tendon, controlled by the linear stepper motor, operated within a range of 0-1440 steps, progressing in 10-step intervals. When the tendon reached its peak displacement, the straightening cycle was initiated. The tendon was gradually retracted in intervals of 10 steps to restore the CDM to its original central position. With each incremental step, a brief stabilization period was implemented, during which a sequence of the Arduino series output was logged and five frames of image-capturing section positions were taken. For each trial, displacement data was acquired from the encoders and bending data from the RFS unit. This procedure was reiterated three times with RFS sliding into five unique CDM sections to verify the repeatability of the data collection procedure.

\par 
Fig. \ref{fig:variable} shows the type of dataset recorded as input to  the deep neural network for model training. First, the number and position of each red color marker is detected by the camera as $N$, $P(x,y)$ and $P'(x,y)$ represents the position of each marker at a bending configuration. $\Delta E_L$ and $\Delta E_R$ are the offset between the original RFS tip position without bending and the current tip position after bending, as recorded by the encoders; $E$ is a fixed distance between each color mark measured by the encoders; and the final value of  $E_L$ and $E_R$ are the sum of $E$ and corresponding $\Delta{E_L}$ and $\Delta{E_R}$. We insert and extract the RFS unit inside the CDM by $E$ to estimate the bending angles located at these markers. Based on these joint angles, we can determine the bending of the whole CDM. The values of RFS units were recorded as $R_L$ and $R_R$ indicating the bending of a particular joint of the CDM under that particular configuration.

\subsection{Data Processing}

\par Throughout each trial, a brief stabilizing period was introduced after each incremental step. During this phase, data from the RFS units and encoders were captured and subsequently averaged to mitigate noise and enhance data fidelity, especially to eliminate errors caused by RFS drifting and ensure the temporal consistency of the data. From the sequence of five image frames, the positions of all distinct sections were extracted, as well as the position and orientation of the base point in the camera frame. This information facilitated the transformation from the camera frame to the tool frame. Each trial's results were cataloged in a CSV file with dimensions $540\times 14$. In this setup, every one of the 540 columns consisted of metadata combining encoder readings, RFS readings from the corresponding section, and the transformed Cartesian positions of the five sections. In total, the study generated 15 CSV files, collectively forming a robust dataset comprising 8100 data entries. This dataset subsequently served as input for training of a deep learning model.
\begin{figure}[H]
   \centering
    \includegraphics[width=0.46\textwidth]{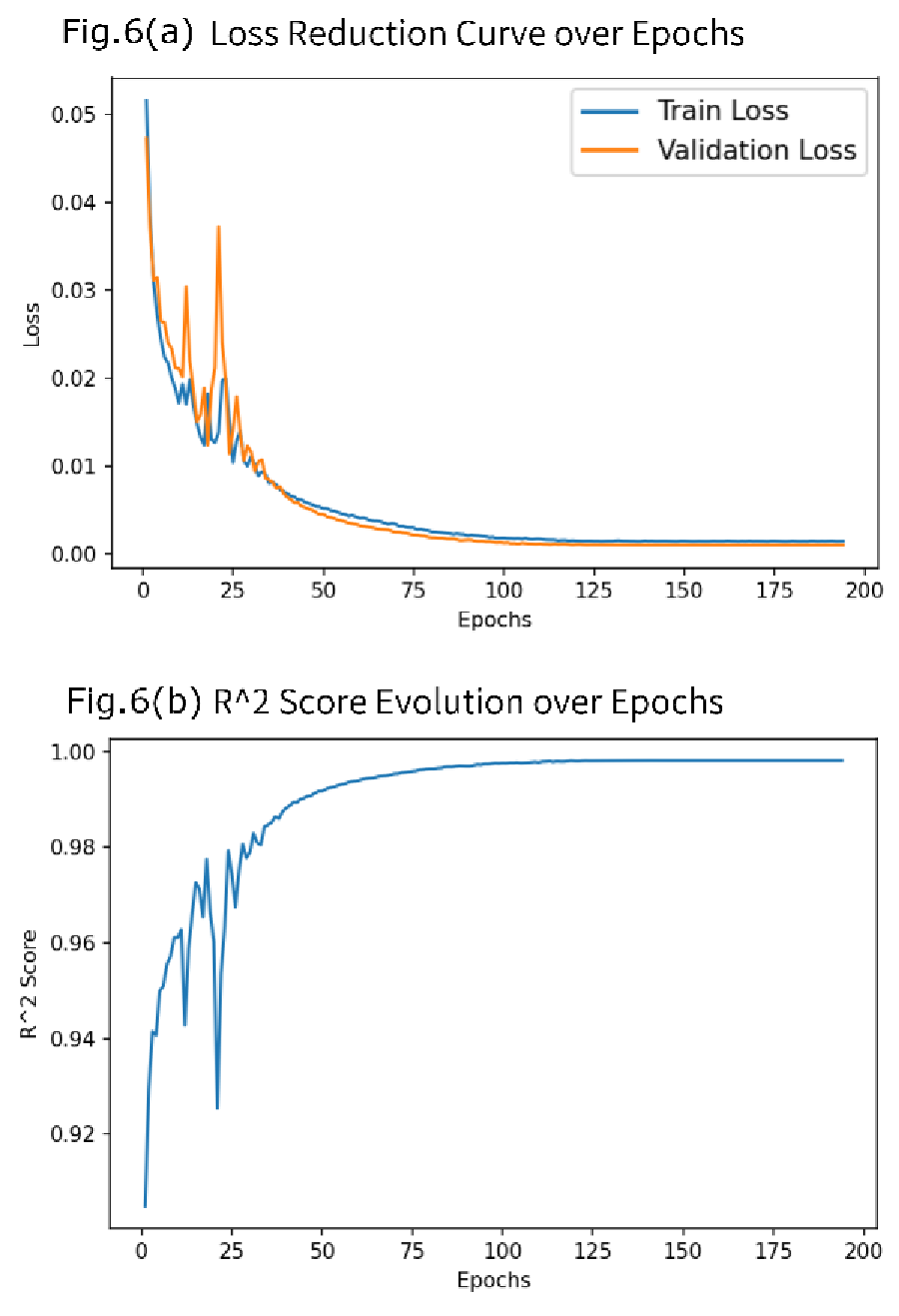}
    \caption{ResNet training and validation loss.}
    \label{fig:loss}
\end{figure}
\subsection{Network Training and Evaluation}

\par Using the generated data set, we developed a deep-learning model referred to Residual Network (ResNet). We used Mean Squared Error (MSE) as loss function for Evaluation and Adaptive Moment Estimation for model optimization. 
\par The learning model has four layers. The first fully connected linear layer expands the initial four input features ($R_L$, $R_R$, $E_L$, $E_R$) into a broader space of 1024 features. Then Two Residual blocks follow the initial expansion. Finally, the last fully connected linear layer primarily maps the 1024 features, processed and refined through the hidden layer, into two outputs labeled as $N$ and $P(x,y)$. In Fig. 6(a), the training and validation losses show initial noise, indicating potential erroneous features. As training advances, these are rectified, with both losses stabilizing at a value of 0.0019. This pattern signifies effective learning and the model's adeptness at capturing data patterns. The tight alignment of the training and validation curves demonstrates the model's robustness and generalization. In addition, the coefficient of determination \( R^2 \) in Fig. 6(b), quantifies how well a model's prediction match the actual data. The curve shows initial fluctuations, but the value impressively stabilizes at 0.998, indicating our model accounts for 99.8\% of the variance, highlighting its strong predictive accuracy.
\subsection{Results}

\begin{figure}[H]
   \centering
    \includegraphics[width=0.49\textwidth]{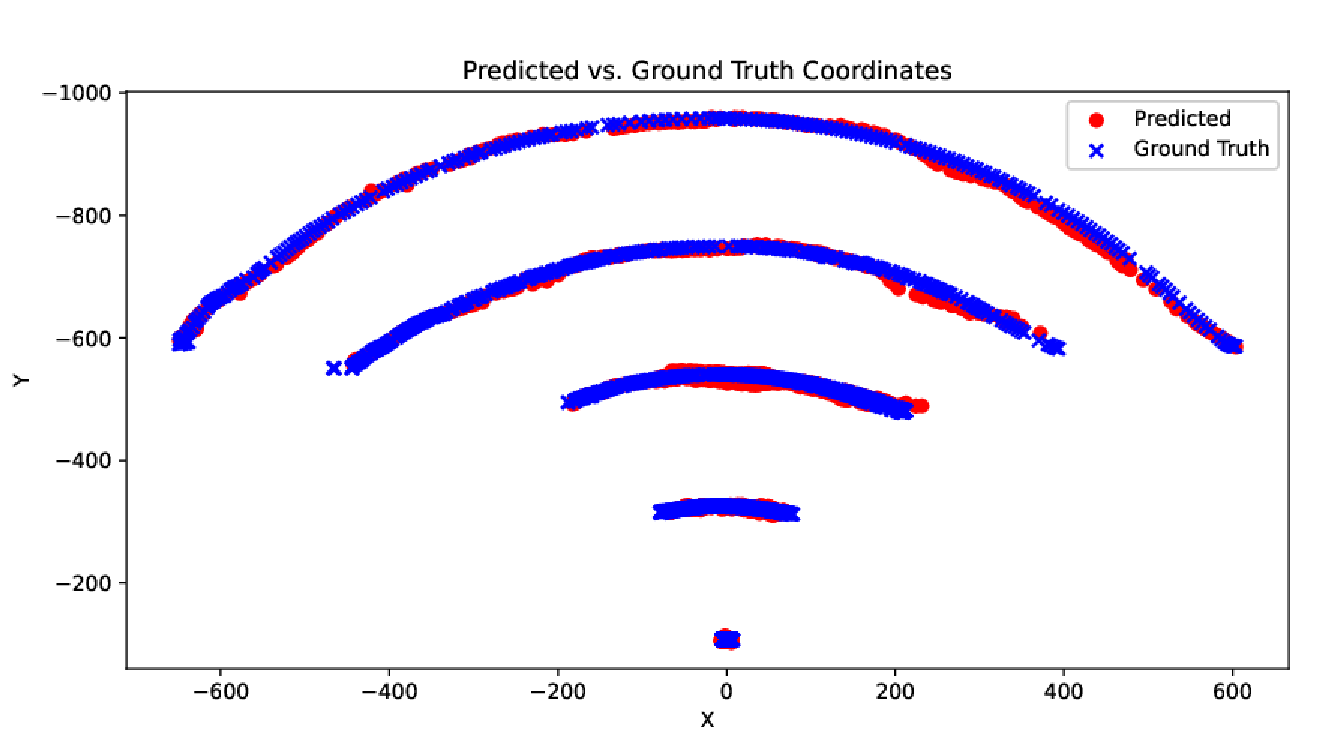}
    \caption{RFS scanning configuration space within the CDM.}
    \label{fig:train}
\end{figure}

\begin{figure}[H]
   \centering
    \includegraphics[width=0.49\textwidth]{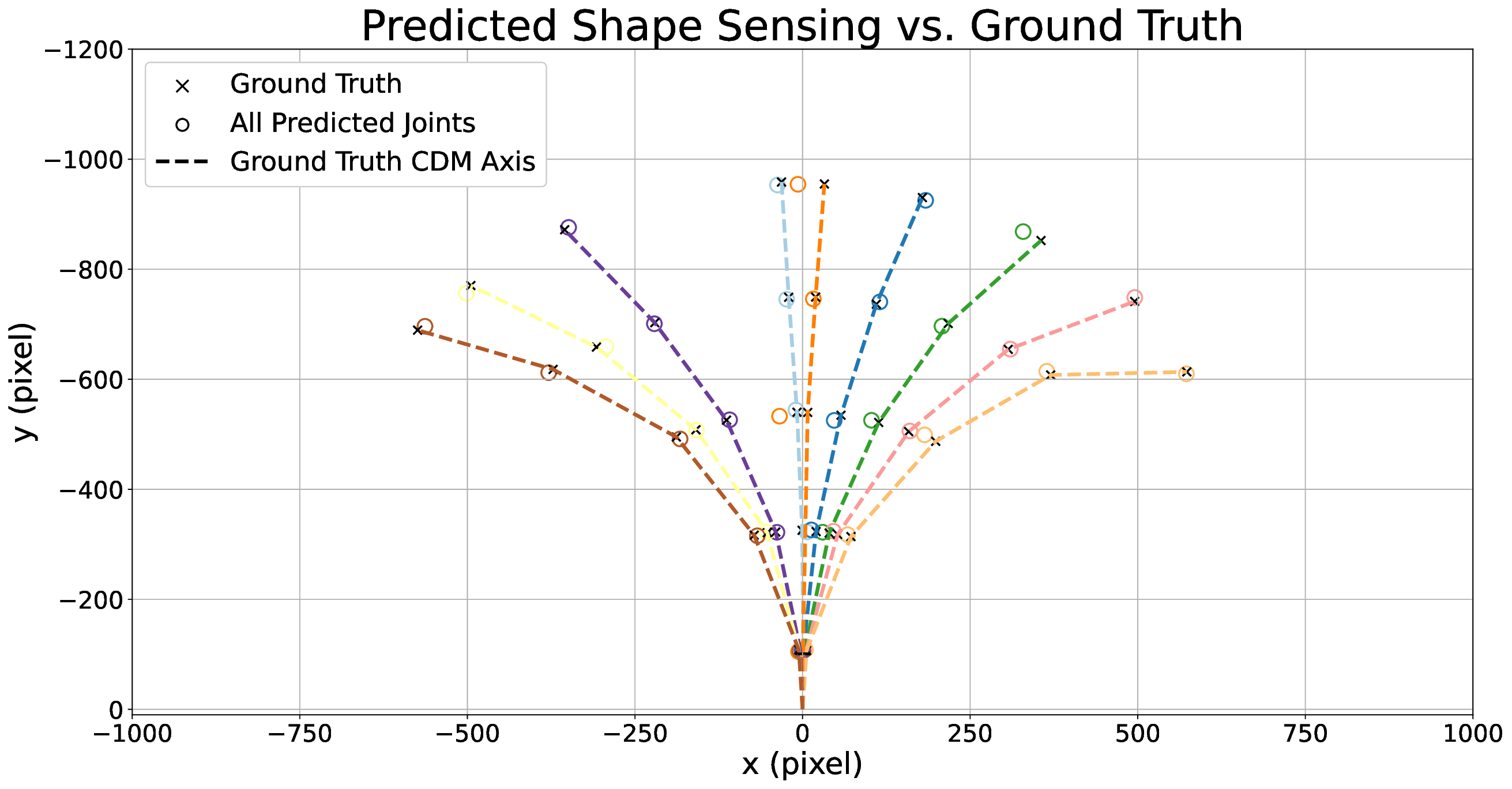}
    \caption{CDM 2D reconstruction after RFS calibration.}
    \label{fig:recons}
\end{figure}
\begin{table}[H]
\caption{Joint Errors Comparison} 
\setlength{\defaultaddspace}{8pt} 
\setlength{\tabcolsep}{5pt} 
\begin{tabular}{lcccccl}
\toprule[.1em] 
Joint number      & 1     & 2     & 3     & 4     & 5     & Total \\ 
\midrule
Average err (mm) & 0.238 & 1.07 & 1.26 & 1.25 & 1.43 & 0.968 \\ 
\midrule
Standard err (mm) & 0.0688 & 0.414 & 0.735 & 0.939 & 0.793 & 0.275\\
\bottomrule[.1em] 
\end{tabular}
\end{table}
The data presented in Table I highlights the prediction discrepancies of the model concerning the positions of various joints.  These joints are categorized from 1 to 5 based on their proximity to the CDM base, with joint 1 being the most proximal and joint 5 the most distal.  Upon examination, it's evident that the errors exhibit a pattern of incremental increase from the base towards the tip.  Nonetheless, the overall precision remains acceptable.  Our analysis suggests that the minimal error observed at joint 1 can be attributed to its constrained deformation space, coupled with the fact that RFS is less responsive to features with a diminutive radius of curvature. The increased  error at the tip may be due to its  larger deformation space.  Furthermore, while the introduction of the pre-tension mechanism has significantly reduced hysteresis, this phenomenon is notably more pronounced at the tip.

Fig.  \ref{fig:train}  shows the RFS as CDM is bended +/- 90 degrees. 
The figure shows that the predicted RFS scans closely correspond to the data measured by the overhead cameras (ground truth). As is shown in Fig. \ref{fig:recons},  the reconstruction of the CDM bending from the camera views closely follows the reconstruction of the joint positions based on the neural network model.


\subsection{Discussion}

\par In this work, we embedded a resistive flexible sensor (RFS) to detect the shape of a cable-driven CDM in real-time. The main goal is to investigate an inexpensive, accurate, and robust shape sensing technique to provide CDMs shape feedback during minimally invasive surgical procedures. This alternative approach reduces the cost of shape sensing in continuum manipulators by two orders of magnitude when compared to the use of FBG sensors that are commonly used in needle steering and are proposed for CDM shape sensing. Such cost reduction is especially important for transitioning the use of CDMs to minimally-invasive applications since CDMs have a limited life cycle and will need to be relatively inexpensive. Moreover, in contrast to the FBG sensors, RFS is not sensitive to the environmental changes of the temperate when CDM is used for drilling and milling of hard tissues such as bone. Furthermore, the presented shape sampling method enables non-constant curvature configurations. In the future, we will train and show the applicability of the RFS unit for accurate tracking of the CDM when it assumes more complex  (e.g. S-shape) configurations due to external loads and other environmental conditions.

\par This alternative approach reduces the cost of shape sensing in continuum manipulators by two orders of magnitude when compared to the use of FBG sensors that are  commonly used in needle steering and are proposed for the CDM shape sensing. Such cost reduction is especially important for transitioning the use of CDMs to minimally-invasive applications, since CDMs have limited life cycle and will need to be relatively inexpensive. Moreover, in contrast to the the FBG sensors, RFS is not sensitive to the environmental changes of the temperate when CDM is used for drilling and milling of  hard tissues  such as bone. 

\par The RFS scanning unit spans an active length of 95.25mm, which adequately covers the entire length and +/- 90 degrees of bending of the CDM.  This effectively covers a wide range of the CDM bending in two directions.  By placing two opposing RFS units on each side of the CDM, we ensure full coverage of the 2D configuration of the CDM from -90 degrees to +90 degrees. Furthermore, the encoders of the scanning unit continuously monitor the sensor's insertion depth into the CDM. Currently, our experiment focuses on scanning sensors in five distinct CDM joints. However, this design also offers sufficient flexibility for continuous scanning across all joints, potentially improving the overall accuracy of shape prediction, especially when assuming more complex shapes such as S-shape.

\par We implemented sensor parameter training for calibration using a deep learning network, ResNet. The vector of inputs were expanded to a larger space of features, and the features were subsequently refined by residual blocks resulting in a robust model. Furthermore, the $R^2$ value of 99.8$\%$ demonstrates the precision of the predictions. Thus, combining traditional sensing with advanced machine learning holds great promise for medical tasks that require high precision.

\par While the methodology showed promise, our approach still had several limitations. Our current technique scans multiple discrete joint positions to estimate the complex global shape of the CDM. This introduces a margin of error, especially for unmeasured joints. A potential venue for future research involves exploring fully automated motor-driven RFS units with the goal of enhancing the scanning workflow and expediting the sensing procedure. Additionally, the ResNet model employed for sensor calibration could potentially be adapted to accommodate sequential dynamic input,  which would facilitate real-time adjustments to shape alterations. Further research into miniaturizing the RFS unit would allow it to be used in smaller continuum manipulators, offering greater potential in minimally invasive surgery.

\section{Conclusion}

In summary, our study aimed to develop, and validate a novel shape reconstruction technique embedded in an Continuum Dexterous Manipulator (CDM) equipped with RFS-based scanning unit. Furthermore, incorporating the ResNet machine learning model into our approach allowed the sensor to be easily calibrated without a mathematical model. Through our empirical tests, the model has demonstrated its ability to predict the non-uniform configurations of the CDM in real-time. Our results confirmed the viability of the proposed data-driven sensor scanning model for CDM shape sensing in non-constant curvature bending scenarios.


\printbibliography












\end{document}